\documentclass[10pt,twocolumn,letterpaper]{article}

\usepackage[pagenumbers]{cvpr}      

\definecolor{cvprblue}{rgb}{0.21,0.49,0.74}
\usepackage[pagebackref,breaklinks,colorlinks,allcolors=cvprblue]{hyperref}

\def\paperID{6286} 
\def\confName{CVPR}
\def\confYear{2025}

\title{Manga Generation via Layout-controllable Diffusion}

\author{Siyu Chen$^{*}$, Dengjie Li$^{*}$, Zenghao Bao, Yao Zhou, Lingfeng Tan, Yujie Zhong, Zheng Zhao\\
Meituan Inc.\\
{\tt\small \{chensiyu25,lidengjie,baozenghao,zhouyao11,tanlingfeng,zhongyujie,zhaozheng08\}@meituan.com}
}

\usepackage{hyperref}
\usepackage{url}
\usepackage{stfloats}

\begin{document}
\maketitle

\renewcommand{\thefootnote}{}
\footnote{$^{*}$Equal contribution}

\begin{abstract}
Generating comics through text is widely studied. However, there are few studies on generating multi-panel Manga (Japanese comics) solely based on plain text. Japanese manga contains multiple panels on a single page, with characteristics such as coherence in storytelling, reasonable and diverse page layouts, consistency in characters, and semantic correspondence between panel drawings and panel scripts. Therefore, generating manga poses a significant challenge. This paper presents the manga generation task and constructs the Manga109Story dataset for studying manga generation solely from plain text. Additionally, we propose MangaDiffusion to facilitate the intra-panel and inter-panel information interaction during the manga generation process. The results show that our method particularly ensures the number of panels, reasonable and diverse page layouts. Based on our approach, there is potential to converting a large amount of textual stories into more engaging manga readings, leading to significant application prospects. Project page: \url{https://siyuch-fdu.github.io/MangaDiffusion}.

\end{abstract}

\section{Introduction}
\label{sec:intro}

\begin{figure}
\setlength{\abovecaptionskip}{0cm}
\begin{center}
    \includegraphics[width=1\linewidth]{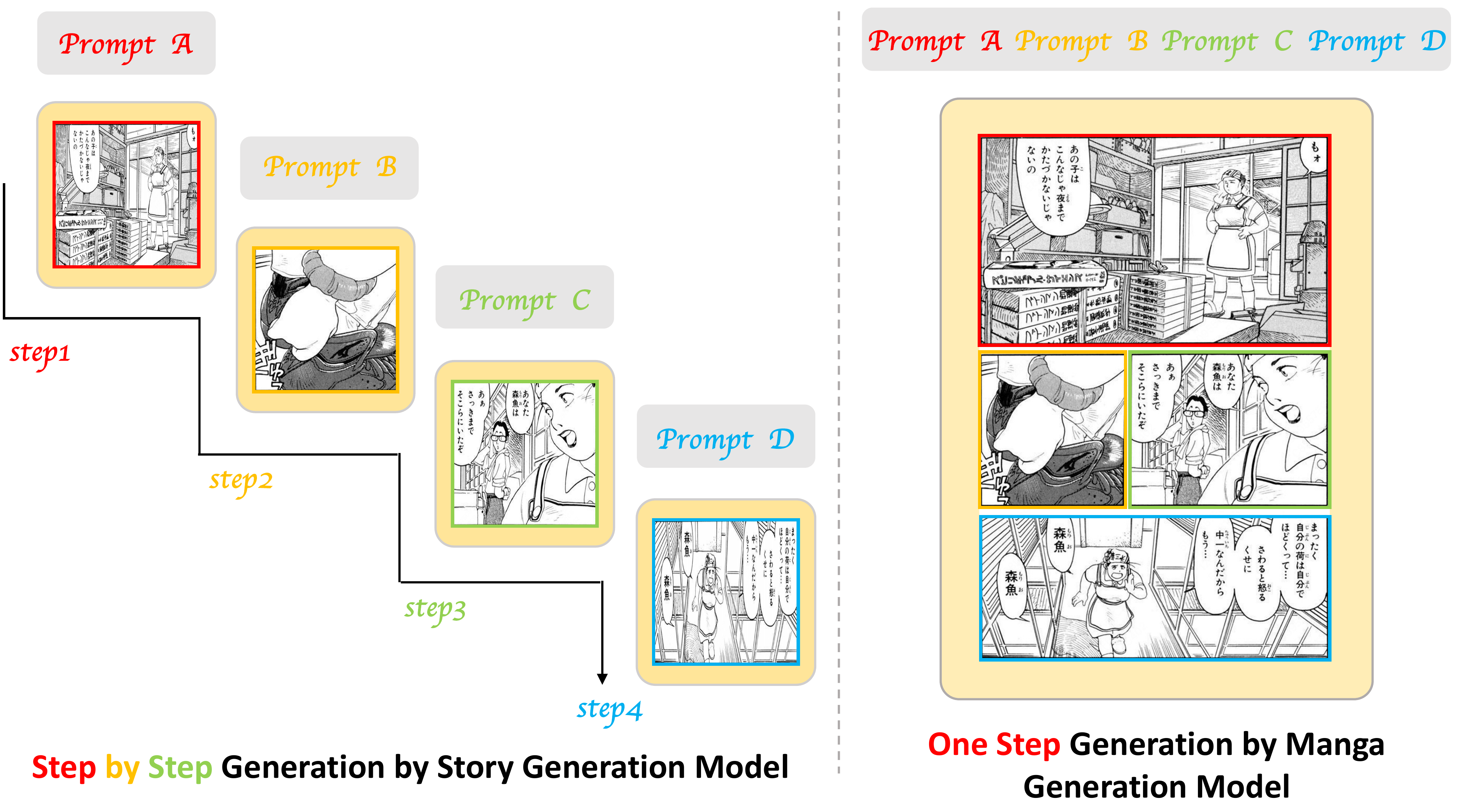}
\end{center}
   \caption{Difference between story generation task and manga generation task. (a) The prompt is inputted into the model one by one. The generation of images is controlled independently, therefore it is no need to consider the reading order and layout of the images. (b) The prompt is inputted into the model as batches, and the model processes all inputs at once. Each input controls the content generation of each panel, and the model controls the overall order and layout of the panels, ultimately outputting a manga page with multiple panels.}
\label{fig:motivation}
\end{figure}

The core goal of an information carrier is to convey semantics, and visual information plays a crucial role in it as a unique modality. As the saying goes, \textit{a picture is worth a thousand words.} Nowadays, various forms of visual arts are based on this characteristic. Comics, as an art form that has remained popular for a long time, have the unique ability to express stories through engaging visual images. An excellent comic can convey a memorable story with minimal textual information. Generative models have recently demonstrated impressive capabilities in generating images from text. More and more research is focusing on how to generate comics with continuous stories. So far, many works have focused on the study of \textit{story generation}, which involves generating a series of images based on a series of texts to visualize the story (See Figure \ref{fig:motivation} (a)). This paper focuses on a novel and yet challenging scenario: \textit{manga generation}. The goal of manga generation is to generate multi-panel manga page solely based on the textual story (See Figure \ref{fig:motivation} (b)). Each of these individual panels need to satisfy the requirements of coherent storytelling (\textit{i.e.}, the order of the panels should follow the reading order), reasonable layout, consistency of characters, and semantic correspondence between panels and the narrative of the story. Manga generation has considerable potential in design, business, and other fields. It can assist comic artists in designing manga scripts and help animators in designing storyboard images and layouts. It has the potential to converting a massive amount of plain textual content into more engaging manga materials, thus bringing significant commercial prospects.

In order to achieve manga generation, the first requirement is to build a manga caption dataset. We construct the Manga109Story dataset by using the existing Manga109 dataset\cite{matsui2017sketch}, which is widely used in the field of image super-resolution. Manga109 is currently the largest open-source manga dataset and it offers rich annotation resources that can be utilized, including the coordinates of panels, characters, texts, and the contents of dialogues. These information are crucial for understanding the elements within the panels. Furthermore, many prominent works, such as Manga109Dialog \cite{li2024manga109dialog} dataset and panel order estimator algorithm \cite{7351614}, can further help us understand the narrative information conveyed by manga. Building on these works, we have integrated this information and used the powerful comprehension capability of Multimodal Large Language Model (MLLM) to generate plot-relevant captions to each panel and summarize a story that aligns with the content of the manga page. We have named this dataset Manga109Story and we hope that this dataset can assist the academic community in conducting better research on manga generation.

To generate manga based on plain text, we divided the generation process into two main steps: (1) using LLM to split the complete plain text story into multiple scripts; (2) inputting the scripts into a generation model to generate manga pages end-to-end. In the second step, we propose the MangaDiffusion model, which splits a manga page into multiple panel images. The visual content of each panel image is controlled by the corresponding script, and the number of panel images is controlled to ensure consistency with the number of script. Additionally, intra-panel transformer blocks are designed to enable information interaction within panel images, while inter-panel transformer blocks facilitate information interaction between panel images, ensuring both a reasonable and flexible layout among panel images. Furthermore, in the generation process, to remove the influence of text and speech bubbles on the generated visuals, we utilize a speech bubble segmentation algorithm to identify all speech bubbles. During the training phase, we mask the regions containing speech bubbles, excluding the pixels within these regions from loss calculation, effectively reducing the clutter caused by speech bubbles in the generated visuals.



In summary, the contributions of this paper are as follows: 
\begin{itemize}
\setlength{\itemsep}{2pt}
\setlength{\parsep}{2pt}
\setlength{\parskip}{2pt}
\item 
Introducing the manga generation task, a challenging task aiming to generate manga with multiple panel scenes on a single page based on a plain text story. It requires maintaining continuity between panels, consistency in character depictions, reasonable layout, and correspondence between manga and story.

\item 
Constructing the Manga109Story dataset based on the existing Manga109 dataset. We use MLLM to generate caption for each manga panels and summarize a story for each page. It provides data materials for subsequent manga generation.

\item 
Introducing the MangaDiffusion method, which allows interaction and perception of information among panels during the generation process.

\item Proposing a local mask strategy to reduce speech bubble generation and improve the quality of generated manga.

\end{itemize}

\begin{figure*}
\setlength{\abovecaptionskip}{0cm}
\begin{center}
    \includegraphics[width=1\linewidth]{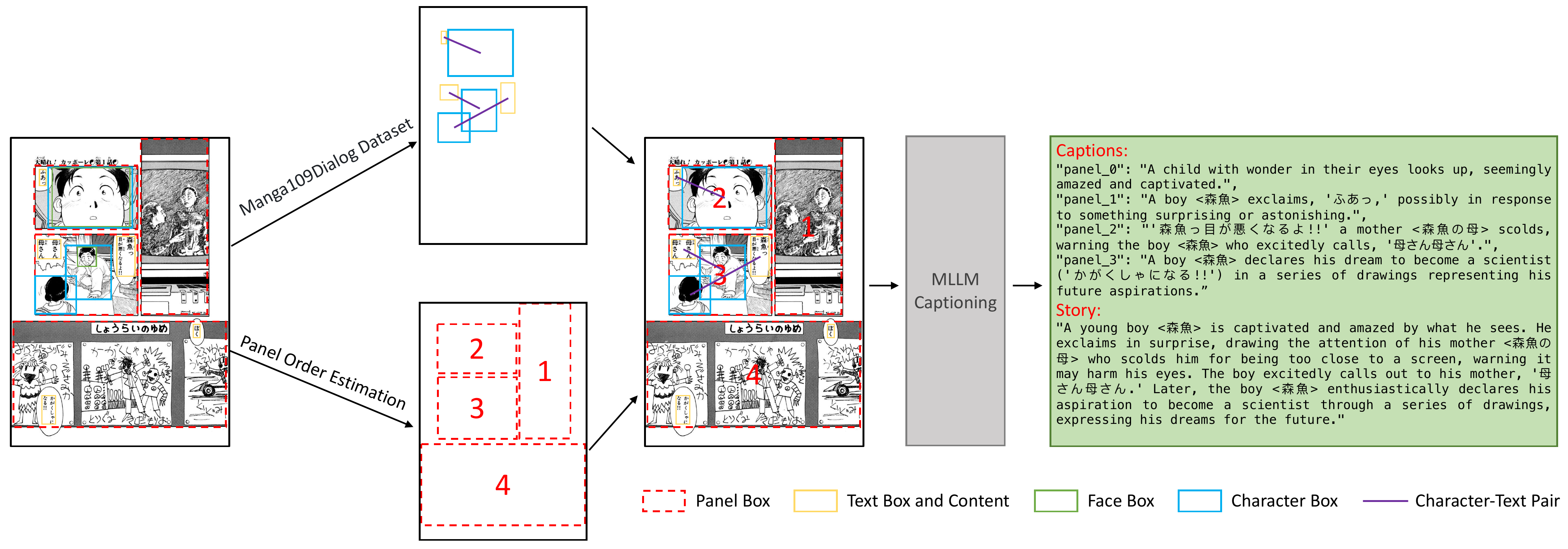}
\end{center}
   \caption{The construction process of Manga109Story dataset. The Manga109 dataset includes basic information such as coordinates of panel, character, face, and text, as well as the text content. The Manga109Dialog dataset associates dialogues with their respective speakers. We utilize a panel order estimator to predict the panel order of each manga page. By combining this information, we create an XML file and input it together with the original manga page into the MLLM for captioning. Ultimately, we obtain captions for each panel and summarize the entire page with a story.}
\label{fig:dataset}
\end{figure*}

\section{Related Work}
\label{sec:related_work}

\paragraph{Text-to-Image Generation.}
Text conditioning image generation is a widely studied task. In the early days, most of the work focused on generating high-quality images using Generative Adversarial Networks (GANs) \cite{goodfellow2020generative}, with representative works including \cite{xu2018attngan, zhang2017stackgan, zhang2018stackgan++}. Auto-regressive transformers, such as DALL-E \cite{ramesh2021zero} and CogView \cite{ding2021cogview}, have also achieved significant success. Recently, diffusion models \cite{ho2020denoising} have shown strong controllability and have gained great attention in text-to-image generation. The representative works include Imagen \cite{saharia2022photorealistic} and DALL-E 2 \cite{ramesh2022hierarchical}. The introduction of stable diffusion \cite{rombach2022high} allows the diffusion process to be performed in the latent space, greatly accelerating the efficiency of diffusion model training, leading to the emergence of a large number of high-quality generative models \cite{podell2023sdxl, peebles2023scalable, zhang2023adding}.

\paragraph{Story Generation.}
The story generation tasks can be divided into two categories: story visualization and story continuation, depending on whether reference images are provided. The story visualization task was first proposed in StoryGAN \cite{li2019storygan}, which is a GAN-based model used to generate cartoon images for stories. Since then, there have been many GAN-based works such as \cite{maharana2021integrating,maharana2021improving,chen2022character,li2022word}. The story continuation task was first proposed in StoryDALL-E \cite{maharana2022storydall}, which provides the model with reference images to obtain sufficient appearance information for the characters. AR-LDM \cite{pan2024synthesizing} proposes a auto-regressive latent diffusion model that introduces a sequential image-caption history as a guide for generating the current frame. Similarly, Story-LDM \cite{rahman2023make} uses a auto-regressive latent diffusion model and designs a visual memory module to capture contextual information of the generated frames, such as characters and backgrounds. TaleCrafter \cite{gong2023talecrafter} presents a fine-grained interactive story visualization system that supports layout and local structure editing of targets in a single image. StoryGen \cite{liu2024intelligent} proposes an open-ended visual storytelling task that can generalize to unseen characters without the need for additional optimization.

\paragraph{Manga}
As a unique form of visual artistic creation, manga has a significant reading and creative community. It stands apart from other realistic images, presenting distinct visual symbols and styles. As a result, various fields of research related to manga have emerged. These fields include manga retrieval \cite{matsui2017sketch}, manga specker detection \cite{li2024manga109dialog}, manga text detection \cite{del2020unconstrained}, manga colorization \cite{qu2006manga,sato2014reference,sykora2009lazybrush}, vectorization \cite{kopf2012digital}, layout recognition \cite{tanaka2007layout}, layout generation \cite{cao2012automatic,hoashi2011automatic}, element composition \cite{cao2014look}, manga-like rendering \cite{qu2008richness}, speech balloon detection \cite{rigaud2013active}, segmentation \cite{aramaki2014interactive}, face-tailored features \cite{chu2014line}, screen tone separation \cite{ito2015separation}, and retargeting \cite{matsui2011interactive}. We have also noticed some works focusing on manga generation, but some of them require inputting illustrations to generate manga \cite{zhang2021generating,li2011content,wu2014mangawall,xie2020manga}, while others require sketch input \cite{lin2024sketch2manga,wu2023shading,li2023reference,zhang2018two,zou2019language,zhang2021user,tsubota2019synthesis}. To our knowledge, there is currently no research on directly generating manga from plain text.
\begin{figure*}
\setlength{\abovecaptionskip}{0cm}
\begin{center}
    \includegraphics[width=1\linewidth]{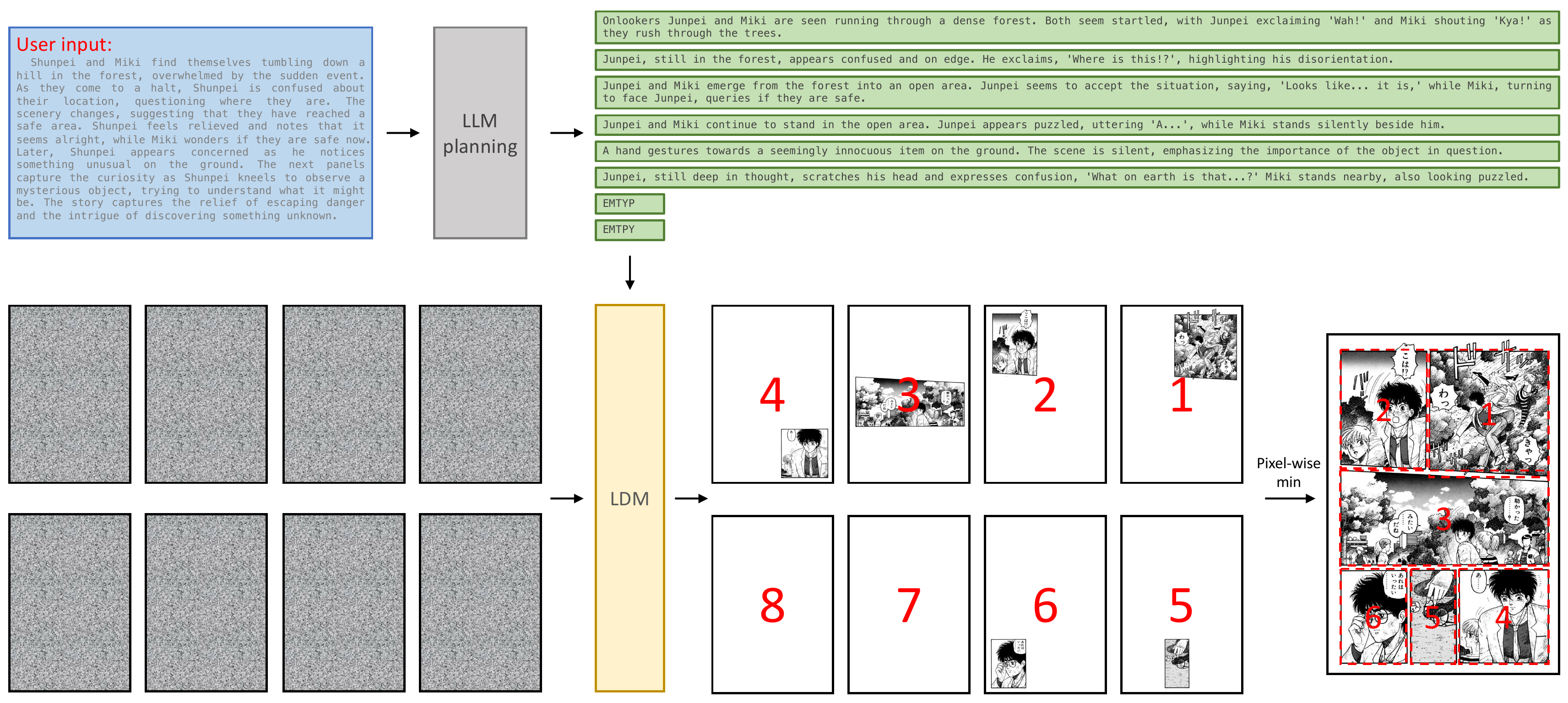}
\end{center}
   \caption{The entire pipeline of our proposed manga generation method. Users input a plain text story, and with the help of LLM, we plan the story and divide it into $K$ scripts. If $K$ is smaller than the maximum supported number of panels, the script will be filled with \textit{EMPTY}. The scripts and randomly sampled Gaussian noise are then fed into our proposed MangaDiffusion model for manga generation, resulting in $K$ ordered panels. By taking the minimum value pixel-wise for each panel, a manga page is synthesized.}
\label{fig:pipeling}
\end{figure*}

\section{Manga109Story Dataset}
\label{sec: manga109story}

In this section, we introduce how to construct the Manga109Story dataset based on existing community work and leveraging the capabilities of multimodal large language model (MLLM).

\paragraph{Existing Works.}
Manga109 is an open-source manga dataset \cite{matsui2017sketch}.
Manga109 consists of 109 volumes of manga drawn by professional manga artists in Japan, totaling 21,142 pages. These manga were commercially distributed to the public from the 1970s to the 2010s, covering a wide range of target readers and genres. This dataset is widely used in the fields of image super-resolution and image restoration. In this dataset, the authors provide rich annotation information, as shown in Figure 2, including the coordinates of the bounding boxes for each frame (referred to as \textit{panel} in this paper), the names and coordinates of each character, the coordinates of each face, and the coordinates and content of each text. These information is the most basic visual elements. With these annotation, we can perform simple captioning on the images. However, there are still difficulties in understanding and describing the storyline depicted in the manga through these visual elements. Fortunately, the prosperous community ecology and excellent researchers of Manga109 have effectively extended the Manga109 dataset.

Manga109Dialog \cite{li2024manga109dialog} is a large-scale dialogue dataset for comics speaker detection. The purpose of this work is to promote machine understanding of comics. The authors associate the dialogues and speakers in Manga109, transforming the originally coordinate-based text data into structured information, indicating which character says a certain sentence. This can help us better understand the storyline depicted in the panels.

Manga109 community offers a panel order estimator which uses the method proposed in \cite{7351614} to predict the order of manga panels. The method takes a series of panel bounding boxes as input and outputs the panel order. Obtaining the reading order of manga panels through this method can help us better understand the overall storyline.

\paragraph{Information Integration}
Fig \ref{fig:dataset} shows the construction process of Manga109Story datset.
With the help of previous works, we can integrate various useful information to form an XML file that contains multiple usable information related to the manga page. The XML file mainly consists of three elements: panel, character, and text. Panels are arranged in order, character is the sub-elements of panel, representing the character contained in the panel, and text is the sub-element of character, representing the dialogue spoken by the corresponding character. Finally, we utilize the powerful multimodal understanding ability of MLLM (GPT-4o \cite{achiam2023gpt}) to fed the original manga page and the corresponding XML file into MLLM, and prompt it to generate corresponding captions for each panel and summarize the corresponding story. In the final Manga109Story dataset, we provide captions for each panel and a story for the entire page. 

\section{MangaDiffusion}
\label{sec: mangadiffusion}


In this section, we introduce the Manga generation task in Section \ref{sec: manga_generation}. Then we present the entire process pipeline in Section \ref{sec: pipeline}. Finally, in Section \ref{sec: architecture}, we elaborate on the architecture of the proposed MangaDiffusion model and its corresponding application optimizations.

\subsection{Manga Generation}
\label{sec: manga_generation}

The main objective of manga generation tasks is to generate a manga page that contains multiple panels based on a given set of panel scripts, where each panel depict a scene corresponding to the panel script. Specifically, given a set of panel scripts $S = [s_1, s_2, ..., s_K]$, a page $\hat{P}$ is generated by manga generation methods, where page $\hat{P}$ is composed of $K$ panels $\left\{\hat{p_1}, \hat{p_2}, ..., \hat{p_k}\right\}$, and each $\hat{p_k}$ corresponds to a region in page $\hat{P}$, referred to as a \textit{panel} in this page.


The main differences between manga generation and other story generation tasks are mainly in three aspects: 1) Most story generation tasks generate a single image based on a paragraph, which only contains one panel and requires multiple forward inferences to generate multiple images. In contrast, manga generation outputs a complete manga page with multiple panels through a single forward inference. 2) Some story generation tasks can generate a single image with multiple panel images based on a paragraph, but they cannot customize the story content corresponding to each panel image nor control the number of panel images. However, manga generation allows customization of the panel script corresponding to each panel image and can control the number of panels to match the number of panel scripts. 3) Story generation tasks do not need to consider the layout relationship between generated images, which means the images only need to be generated in order. However, in manga generation, since multiple panels are integrated into a single page, the layout and reading order between panels need to be considered to ensure a smooth reading experience for the readers.

\subsection{Manga Pipeline}
\label{sec: pipeline}


In order to enable the practical application of manga generation, we have built a more user-friendly pipeline, which includes two steps: story script segmentation and manga generation. This allows users to directly input a complete story in plain text and obtain a single manga page with multiple panels. Figure \ref{fig:pipeling} illustrates the complete reasoning process. Leveraging the powerful language understanding capabilities of LLM, we split the user's plain text into $K$ continuous script segments, which are then inputted into the generation model to control the generation of each panel image. After training, our model is able to learn the layout information between panels, enabling the generation of $K$ panels in a reasonable order while ensuring flexibility in panel layout. Finally, we take the pixel-wise minimum value of these generated panels to obtain a complete manga page.


\subsection{Architecture}
\label{sec: architecture}
\begin{figure*}
\setlength{\abovecaptionskip}{0cm}
\begin{center}
    \includegraphics[width=1\linewidth]{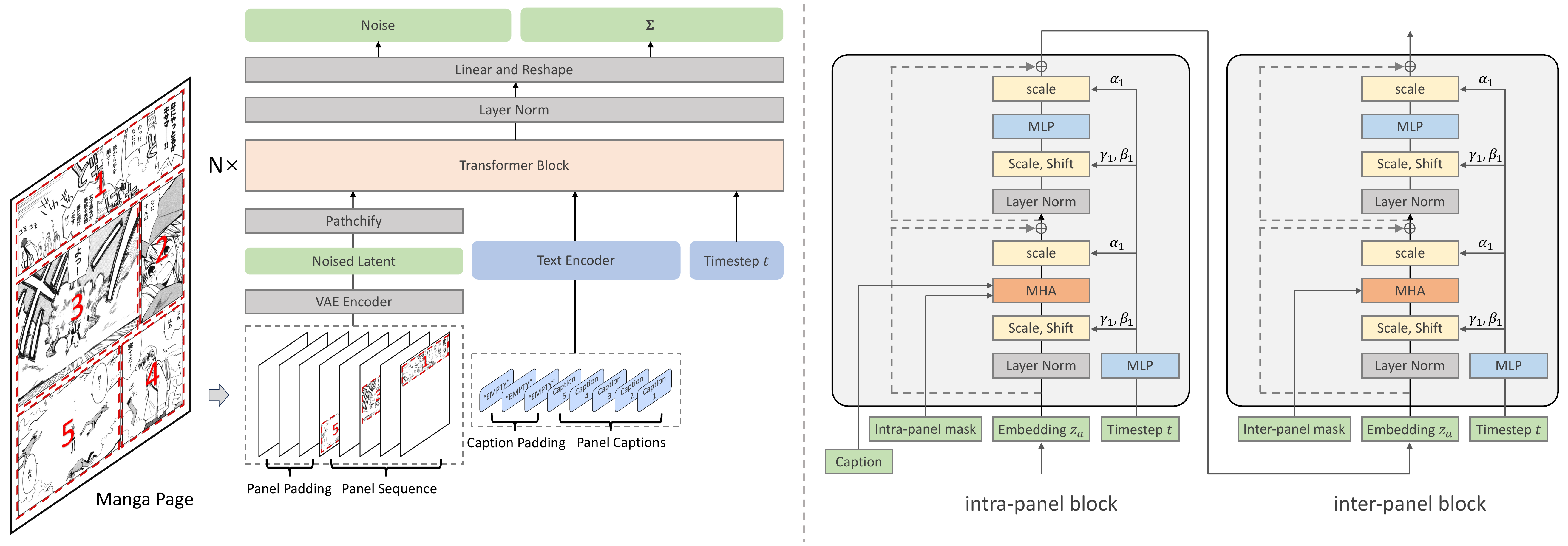}
\end{center}
   \caption{Architecture of MangaDiffusion. During the training stage, we split panel images from a complete manga page. A padding image is added if the number of panels is less than the maximum supported number. These panel images are inputted in batches into a pretrained VAE to obtain the latent representation. Each panel image has a corresponding caption to control its content generation. The transformer block consists of an intra-panel block and an inter-panel block for information interaction. The caption only participates in the computation within the intra-panel block. Timestep $t$ is injected into the model using adaLN-single \cite{chen2023pixart}. The intra-panel mask is used to remove text and speech bubble boxes within the image, while the inter-panel mask is used to mask out the padding images.}
\label{fig:overview}
\end{figure*}

We propose MangaDiffuion to achieve manga generation. Figure \ref{fig:overview} shows the architecture of MangaDiffusion. Specifically, given a manga page $\boldsymbol{P} \in \mathbb{R}^{H \times W \times 3}$ containing $K$ panels, we separate each panel while setting all pixels outside the panels to 1 (white color). This way, we obtain $K$ panel images $\boldsymbol{P_I} \in \mathbb{R}^{K \times H \times W \times 3}$, each panel image with the same size as the original manga page. We use a pre-trained VAE encoder $\mathcal{E}$ to encode each of these $K$ panel images into the latent space, then we obtain $\boldsymbol{P}_{\boldsymbol{L}} \in \mathbb{R}^{K \times 4 \times \frac{H}{8} \times \frac{W}{8}}$. With the patchify process, $\boldsymbol{P}_{\boldsymbol{L}}$ is transformed into input tokens $\boldsymbol{z_I} \in \mathbb{R}^{K \times \frac{H}{16} \times \frac{W}{16} \times d_I}$, where $d_I$ represents the dimension of each token.

As shown in the Figure \ref{fig:overview}, $\boldsymbol{z_I}$ is fed into subsequent Transformer blocks, which are composed of a series of intra-panel blocks and inter-panel blocks. The intra-panel block facilitates feature interaction among all tokens within the same panel, while the inter-panel block facilitates feature interaction among tokens at the same positions across different panels, thereby capturing the layout and corresponding reading order between panels. We reshape $\boldsymbol{z_I} \in \mathbb{R}^{K \times \frac{H}{16} \times \frac{W}{16} \times d_I}$ into $\boldsymbol{z_e} \in \mathbb{R}^{K \times \frac{HW}{256} \times d_I}$ as the input to the intra-panel block, where $\frac{HW}{256}$ corresponds to the number of tokens in each panel. Subsequently, the output of the intra-panel block is transposed to $\boldsymbol{z_a} \in \mathbb{R}^{\frac{HW}{256} \times K \times d_I}$, which serves as the input for learning the layout information in the inter-panel block.


For each caption corresponding to each panel, we obtain the corresponding text embedding through a text encoder $\mathcal{T}$. Specifically, for $K$ captions, after passing through the text encoder, we obtain $\boldsymbol{z_T} \in \mathbb{R}^{K \times l \times d_T}$, where $l$ represents the caption length and $d_T$ represents the dimension of the text embedding. Following adaLN-single \cite{chen2023pixart}, the text embedding $\boldsymbol{z_T}$ is used as the control condition for the intra-panel block and inter-panel block calculation as shown in the Figure \ref{fig:overview}. Captions are used as control conditions only within the intra-panel calculation. Additionally, the $k$-th caption only serves as the control condition for the $k$-th panel. This enables the model to accept multiple captions as inputs and independently control the generation of corresponding panel content. 

The inter-panel mask is used to mask out the padding panels, so that these padding panels do not participate in the attention calculation of the inter-panel block. Through these efforts, the network learns that the padding panels do not contain any visuals and do not participate in the layout. 

\begin{figure}
\setlength{\abovecaptionskip}{0cm}
\begin{center}
    \includegraphics[width=1\linewidth]{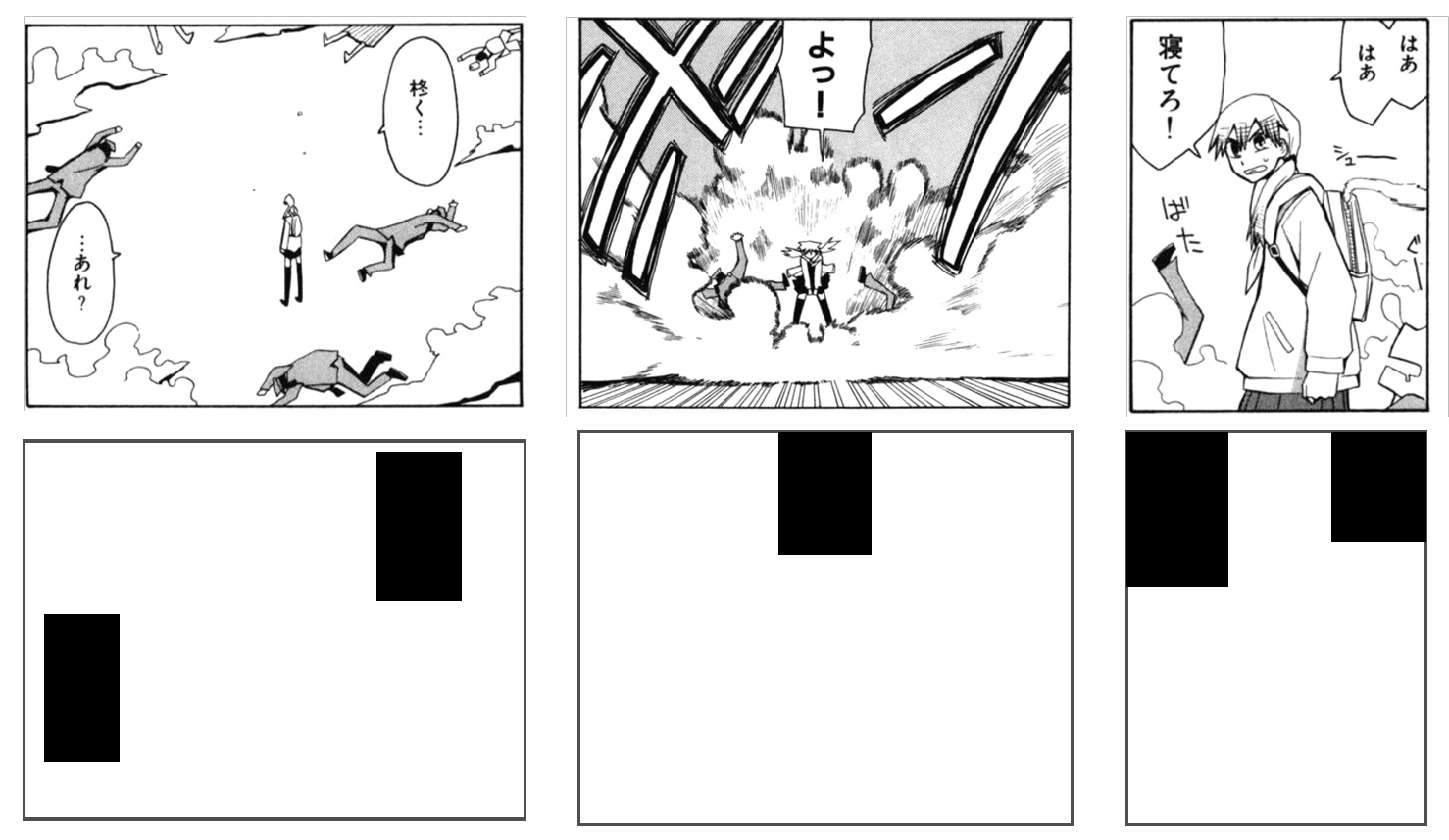}
\end{center}
   \caption{Illustration of intra-panel mask. The first row represents the original panel, and the second row represents the corresponding intra-panel mask. The tokens in the white region are involved in attention calculation and loss calculation, while the black regions are not involved.}
\label{fig:mask}
\end{figure}

The design of the intra-panel mask is aimed at masking out the region of text and bubble box. Figure \ref{fig:mask} shows the illustration of intra-panel mask.
Due to the presence of many character dialogues and corresponding speech bubbles in the original Manga109Story dataset, if this data is used for training without any refinement, the generated data will contain a large number of speech bubbles and garbled characters, which will greatly affect its practical application. Therefore, we utilized an open-source speech bubble segmentation model \cite{bubble_segmentation} and took the maximum bounding box of the segmentation results as a mask to participate in the processing of intra-panel blocks and loss function calculation. We utilize the variational lower bound (ELBO) \cite{kingma2013auto} to minimize the negative log-likelihood of $p_\theta\left(\mathbf{z}_0\right)$ \cite{ho2020denoising}, and the simplified objective can be written as the denoising objective: 

\begin{equation}
    \mathcal{L}=\left\{\begin{array}{lr}
\mathbb{E}_{\mathbf{z}_0, \boldsymbol{\epsilon} \sim \mathcal{N}(0,1), t}\left[\left\|\boldsymbol{\epsilon}-\boldsymbol{\epsilon}_\theta\left(\mathbf{z}_t, t\right)\right\|^2\right], & \text { if }\mathbf{z} \notin \mathcal{M} \\
0, & \text { otherwise }
\end{array}\right.
  \label{eq:intra_mask}
\end{equation}

where $\mathbf{z}_0$ represents the latent of real data, $\mathbf{z}_t$ represents latent representation at time $t$, $\epsilon$ represents the noise predicted by the model, and $\mathcal{M}$ represents the region of the intra-panel mask in latent space.
By adding the intra-panel mask during training, the generated results of the model greatly reduced the number of speech bubbles, making it more suitable for real-world application scenarios, as illustrated in the Figure \ref{fig:Visualization}.

During inference, the generated latents are passed through the VAE decoder, resulting in $K$ panels $\boldsymbol{P}_{\boldsymbol{O}} \in \mathbb{R}^{K \times H \times W \times 3}$. The values at corresponding positions are then combined by taking the minimum value, merging the $K$ panels into one manga page with $H$ height and $W$ width, as shown in the figure \ref{fig:pipeling}.
\section{Experiments}
\label{sec: experiments}

In this section, we provide a detailed description of our experimental settings and compare the results of MangaDiffuion with other image generation methods. Additionally, we compare the visualization results of MangaDiffusion and other SOTAs to demonstrate the effectiveness of our proposed approach.

\subsection{Experimental Settings}
\label{settings}
\paragraph{Backbone.}

Our model is modified based on the Latte \cite{ma2024latte} and utilizes the weights of spatial block and temporal block from the Latte model to initialize the weights of the intra-panel block and inter-panel block in MangaDiffusion.

\paragraph{Implementation Details.}
In order to train manga with different panel numbers in parallel and balance training efficiency, we set the maximum number of layouts to $8$. We discard manga pages with more than $8$ panels. For pages with fewer than $8$ panels, we pad them to $8$ panels. The pixels of the padded panels are all set to $1$, and the corresponding caption for these panels is ``\textit{EMPTY}". We adopts the stable diffusion 1.4 VAE as the image encoder, and the text encoder uses the T5 model with a maximum token count set to $300$. We use the AdamW optimizer with a learning rate of $1e-4$ for training. The batch size is set to $8$, and the training is performed on $4$ NVIDIA A100 GPUs, taking a total of $65$ hours for 100,000 steps. To ensure the reading order of panels, we do not apply common data augmentations such as horizontal flipping to the manga pages. All pages are resized to $[512,384]$ for training.

\paragraph{Dataset.}
Due to limited data resources, we selected one title from the Manga109Story as the test set, which contains 163 manga pages. The remaining 108 manga titles, as the training set, includes a total of 20,979 manga pages.

\paragraph{Evaluation Metrics.}
In order to evaluate the results generated by the manga, we adopt two widely used metrics, that is Frechet Inception Distance score (FID) and CLIP image-image similarity (CLIP-I). Since the captions in Manga109Story are long, exceeding the maximum token length of 77 in CLIP, we do not use CLIP text-image similarity (CLIP-T).

\subsection{Quantitative Results}
We selected a few open-source T2I SOTAs for comparison. These models are not fine-tuned and instead directly take the concatenated panel captions as the prompt to generate corresponding comics. Additionally, we used fine-tuned Pixart-$\Sigma$ \cite{chen2024pixart} on Manga109Story for 120 epochs as the counterpart to non-fine-tuned methods. 

Table \ref{tab:results} shows the results of MangaDiffusion and other T2I SOTAs on the Manga109Story test set. From the table, it can be seen that our model performs well in terms of FID and CLIP-I. For the T2I models that have not been fine-tuned, theirbFID and CLIP-I performances are poor. 
These models have significant differences in data domain compared to the manga data, resulting in less ideal performance. However, the images generated by these models have high image quality, mainly due to their extensive training data and time-consuming training process. Our model does not perform as well as the fine-tuned Pixart-$\Sigma$ when the speech bubbles are removed. However, if we keep the bubbles, we can achieve results comparable to Pixart-$\Sigma$. This is mainly because the ground truth (GT) images in the test set also have speech bubbles. By retaining the bubbles, the generated images visually resemble the GT images more closely, resulting in higher metrics.

\begin{table}
\setlength {\belowcaptionskip} {0cm}
  \begin{minipage}{.48\textwidth}
  \centering
  \setlength{\tabcolsep}{10px}
  \begin{tabular}{@{}ccc@{}}
    \toprule
    model & FID $\downarrow$ & CLIP-I $\uparrow$  \\
    \midrule
    FLUX\cite{flux} & 229.6 & 71.0 \\
    SD3-medium \cite{esser2024scaling} & 192.8 & 76.3 \\
    SD3.5-large \cite{esser2024scaling} & 197.8 & 77.0 \\
    Pixart-$\Sigma$ \cite{chen2024pixart} & 334.5 & 55.7 \\
    \midrule
    Pixart-$\Sigma$ (Fine-tuned) \cite{chen2024pixart} & 149.1 & \textbf{84.6} \\
    \midrule
    MangaDiffusion (Ours) & \textbf{143.5} & 83.6 \\
    MangaDiffusion $wo. bubble$ (Ours) & 179.8 & 75.7  \\
    \bottomrule
  \end{tabular}
  \caption{Comparison between T2I SOTAs and our proposed MangaDiffusion on Manga109Story test set. The Pixart-$\Sigma$ \cite{chen2024pixart} is fine-tuned using Manga109Story train set for 120 epochs, and the other SOTAs are inference directly without any fine-tuning.}
  \label{tab:results}
  
  \end{minipage}
\end{table}

\subsection{Qualitative Results}
\begin{figure*}
\setlength{\abovecaptionskip}{0cm}
\begin{center}
    \includegraphics[width=1\linewidth]{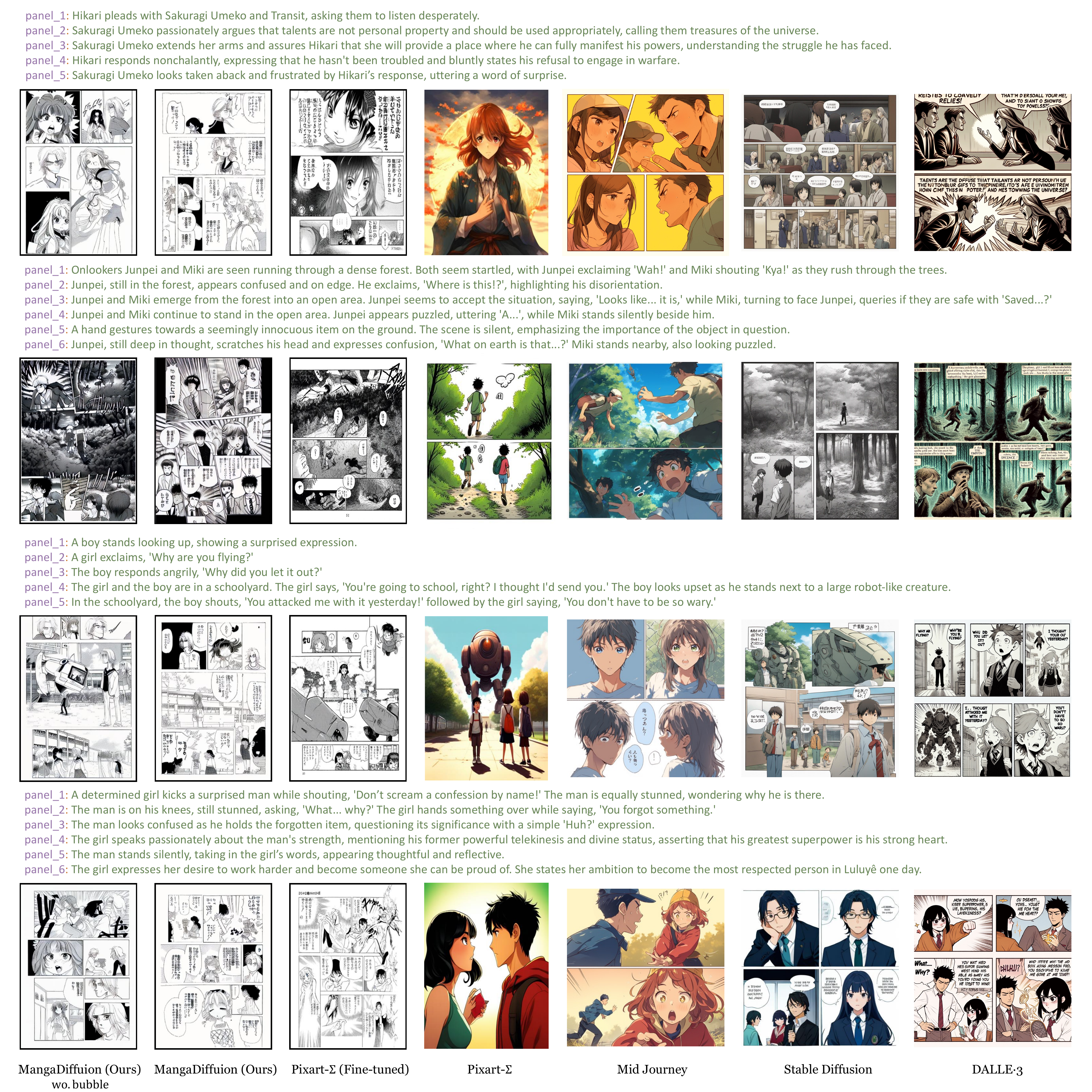} 
\end{center}
   \caption{Visualization results of MangaDiffuion and other T2I methods. Each row represents a story, and the text above the images shows the number of panels and the corresponding captions for each panel. 
   It can be observed from the figures that our method performs well in terms of panel quantity control and panel layout diversity.}
\label{fig:Visualization}
\end{figure*}

Figure \ref{fig:Visualization} shows a comparison of the results of MangaDiffusion and various T2I SOTAs. From the figure, we can see that MangaDiffusion ensures that the number of panels generated in the manga page is consistent with the corresponding prompt input, and each panel depicts a story that has good relevance to the prompt. 
It is worth noting that MangaDiffusion's panel layout is flexible and diverse, not a rigid and uniform layout that is reflected in the results of other methods. At the same time, our method can maintain character consistency between different panels on the same page, much like these compared T2I models, indicating effective information interaction between panels.
Although the T2I models have better image quality, their performance in terms of panel quantity and storyline is not outstanding.


\subsection{Ablation Study}
In order to compare the impact of the speech bubble removal strategy, we conduct ablation experiment on MangaDiffusion test set.
As shown in Table \ref{tab:results}, it can be seen that removing speech bubbles affects the performance, which is mainly because GT also contains bubbles as analyzed above. We would argue that removing the bubbles is more in line with practical application requirements. Dialogue bubbles should only be generated when there are dialogues. Currently, purely relying on data learning without adding any restrictive strategies to generate data with speech bubbles will result in the presence of speech bubbles in scenes without dialogues. However, it is difficult to remove the speech bubbles without affecting the image continuity. On the other hand, it is easy to add text and speech bubbles to panels that do not have speech bubbles. Therefore, training with speech bubble masks is more reasonable.

\section{Discussion}
\label{sec:disscution}
In this section, we discuss the challenges of conducting manga generation under existing resources and the limitations of this paper. 


Data is the foundation for model training. Under the guidance of the scaling law, in order to train a manga generation model with better performance, more training data is undoubtedly required. However, compared with other types of data, collecting manga data is particularly difficult because manga is protected by intellectual property rights, which imposes corresponding restrictions on the collection and use of manga data. The Manga109 dataset used in this paper is the largest open-source dataset known to us. However, the total number of pages is only 21,142, which is far less than other large-scale image datasets. Therefore, we hope that more researchers will be dedicated to constructing larger manga datasets, which can help with pre-training or fine-tuning of manga generation models.


As introduced in Section \ref{sec:intro}, the manga generation task requires not only the rationality of the order and layout between panels but also the consistency of panels with captions and character consistency. The MangaDiffusion we propose currently has great room for development in these two aspects of consistency. 
Our future work will focus on how to improve character consistency, the consistency between panels and captions.


Furthermore, in the field of multi-object generation, current generation models often face the issue of incomplete objectives. The method we propose provides insight into alleviating this problem. Our proposed intra-panel block enables independent control of the generation of different panels within the same image. For multi-object generation scenarios, panels can be considered as objects, and the generation of objects can be controlled through intra-object blocks (equivalent to intra-panel blocks in Sec \ref{sec: architecture}) and inter-object blocks (equivalent to inter-panel blocks in Sec \ref{sec: architecture}). This approach ensures both the fidelity of each individual objective and the improvement of the overall completeness of generated objectives, thus mitigating the problem of incomplete objectives in multi-object generation tasks.

\section{Conclusion}
\label{sec:conclusion}
In this paper, we propose a challenging task: manga generation. Different from the iterative generation of story generation task, manga generation requires generating a manga page with a reasonable panel order, layout, and character consistency based on the story in plain text. To achieve manga generation, we introduce the Manga109Story dataset, which builds upon previous excellent work and employs the multi-modal understanding capability of MLLM to provide detailed captioning for each manga page. Additionally, we propose the MangaDiffusion network, which utilizes transformer blocks for intra-panel and inter-panel information interaction, allowing for flexible and diverse layouts in the generated manga pages while maintaining control over the number of panels. To eliminate cluttered speech bubbles and meaningless characters in the generated manga, we propose a training strategy using local masks to better adapt the generated data to practical applications.
{
    \small
    \bibliographystyle{ieeenat_fullname}
    \bibliography{main}
}

\clearpage
\setcounter{page}{1}
\maketitlesupplementary

\section{Inference Visualization}

\begin{figure*}[!b]
\setlength{\abovecaptionskip}{0cm}
\begin{center}
    \includegraphics[width=1\linewidth]{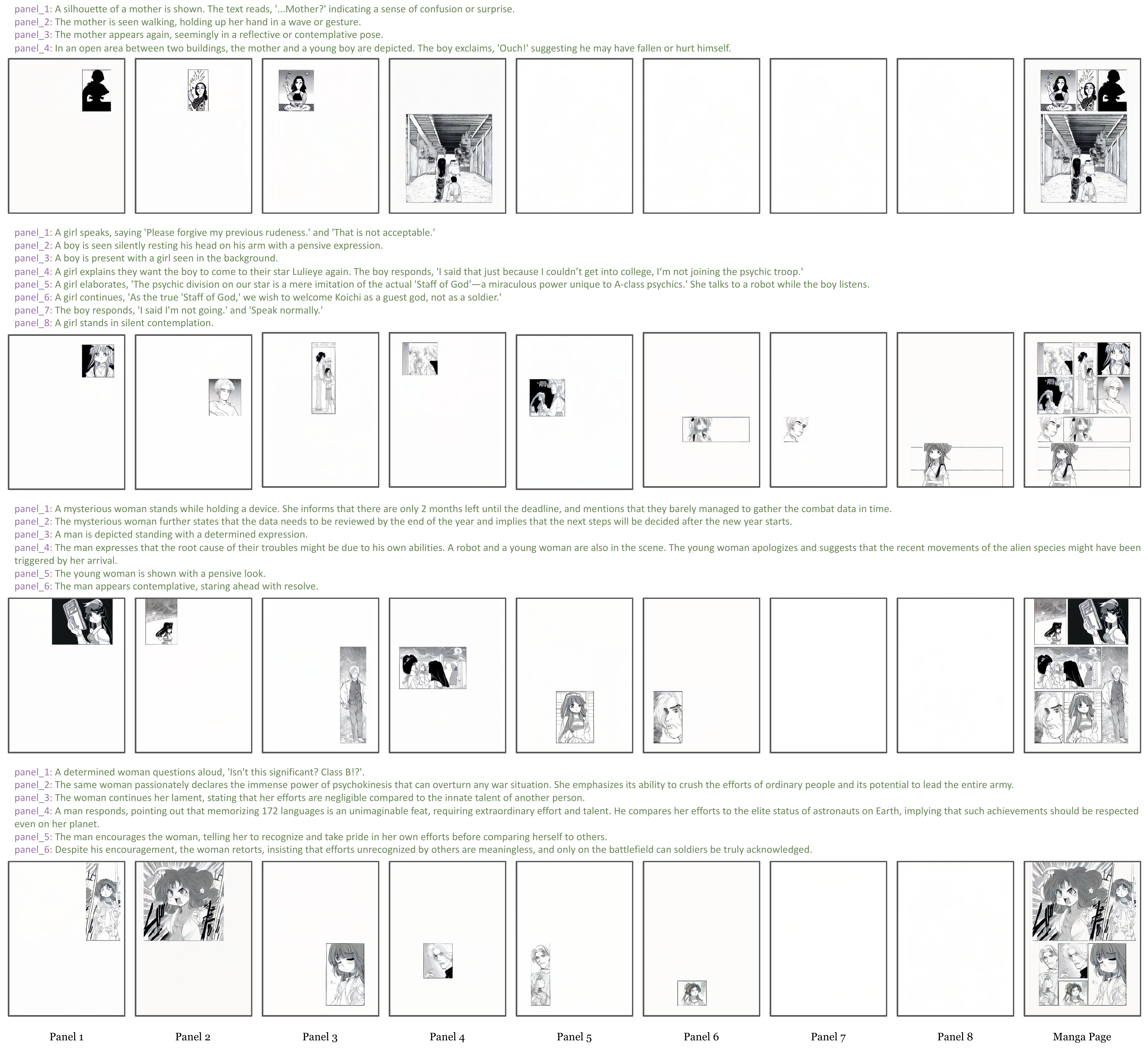} 
\end{center}
   \caption{The visualization of panels and the corresponding manga page. We can see that the arrangement of the content of each panel is distributed in reading order, indicating that our model has learned a reasonable layout. Additionally, in some relatively complex layout scenarios, such as when there is overlapping between panels (the first two panels in row 4), it can be observed that the overlapped content between panels remains almost identical. This also demonstrates that our method enables effective information interaction between panels.}
\label{fig:panels}
\end{figure*}

\vspace{-0.5cm}
Figure \ref{fig:panels} displays the visualization of the inference results, namely, the panels generated by our model, along with the corresponding manga page, which is synthesized by taking the pixel-wise minimum value of each panel. We observe that the contents of the overlapping regions between panels are almost identical (see row 4 in Figure \ref{fig:panels}). This suggests that the inter-panel block effectively facilitates the information interaction between panels, thereby presenting the precise complementarity in the visual contents of each panel.

\section{Manga109Story Construction Details}
\label{sec:detail}

\begin{figure*}[!b]
\setlength{\abovecaptionskip}{0cm}
\begin{center}
    \includegraphics[width=1\linewidth]{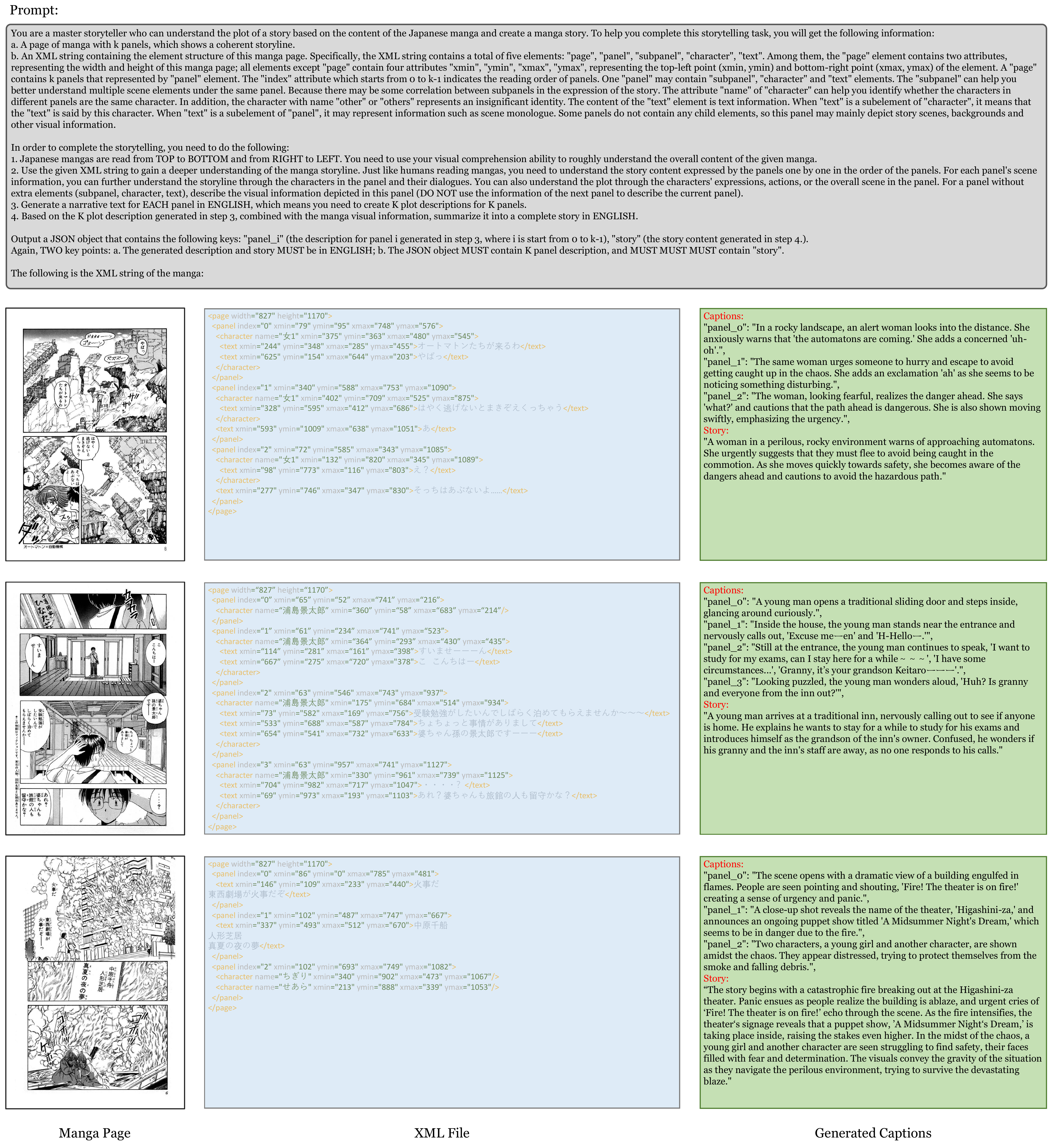} 
\end{center}
   \caption{The details of Manga109Story construction process. The presented XML file is synthesized by integrating the data from the Manga109 dataset, the Manga109Dialog dataset, and the results from the panel order estimator. By inputting the original manga page and XML file into the MLLM, and following the prompts, captions for each panel and a summarized story are generated.}
\label{fig:detail}
\end{figure*}

\vspace{-0.2cm}
To provide a clearer demonstration of the construction process of the Manga109Story dataset, we present a detailed depiction in Figure \ref{fig:detail} of the XML file described in Section \ref{sec: manga109story}, along with the corresponding MLLM prompt and the generated examples.

\end{document}